\documentclass[sigconf]{acmart}
\usepackage{multicol}  
\usepackage{multirow}
\usepackage{booktabs}
\usepackage{subfigure}
\usepackage{amsmath}
\usepackage{caption}
\usepackage[linesnumbered, ruled]{algorithm2e}
\hypersetup{draft}
\usepackage{enumitem}
\usepackage{bbm}
\DeclareMathOperator*{\argmax}{arg\,max}
\DeclareMathOperator*{\argmin}{arg\,min}
\newcommand{\suhang}[1]{\textcolor{blue}{#1}}

\newcommand{\ourModelName}{{NIPA}\xspace}

\begin{document}
\title{Node Injection Attacks on Graphs via Reinforcement Learning}

\begin{abstract}
    Real-world graph applications, such as advertisements and product recommendations make profits based on accurately classify the label of the nodes. However, in such scenarios, there are high incentives for the adversaries to attack such graph to reduce the node classification performance. Previous work on graph adversarial attacks focus on modifying existing graph structures, which is infeasible in most real-world applications. In contrast, it is more practical to inject adversarial nodes into existing graphs, which can also potentially reduce the performance of the classifier.

In this paper, we study the novel node injection poisoning attacks problem which aims to poison the graph.  We describe a reinforcement learning based method, namely NIPA, to sequentially modify the adversarial information of the injected nodes. We report the results of experiments using several benchmark data sets that show the superior performance of the proposed method \ourModelName, relative to the existing state-of-the-art methods.
\end{abstract}


\author{Yiwei Sun, \hspace{0.2em}, Suhang Wang$^{\mathsection}$ ,\hspace{0.2em}Xianfeng Tang, \hspace{0.2em}Tsung-Yu Hsieh, \hspace{0.2em}Vasant Honavar}\thanks{$^{\mathsection}$Corresponding Author}
\affiliation{%
 \institution{Pennsylvania State University}
}
\affiliation{%
  \institution{\{yus162,\hspace{0.2em}szw494,\hspace{0.2em}xut10\hspace{0.2em},tuh45\hspace{0.2em},vuh14\}@psu.edu}
}
\maketitle
\section{Introduction}
Graphs, in which nodes and their attributes denote real-world entities (e.g., individuals) and links encode different types of relationships (e.g., friendship) between entities, are ubiquitous in many domains, such as social networks, electronic commerce, politics, counter-terrorism, among others. Many real-world applications e.g.,
targeting advertisements and product recommendations, rely on accurate methods for node classification~\cite{aggarwal2011introduction, bhagat2011node}. However, in high-stakes scenarios, such as political campaigns and e-commerce, there are significant political, financial, or other incentives for adversaries to attack such graphs to achieve their goals. For example, political adversaries may want to  propagate fake news in social medias to damage an opponent's electoral prospects~\cite{allcott2017social}. The success of such attack depends on a large extent of the adversaries' ability to misclassify the graph classifier.

\begin{figure}[ht]
    \centering
    \includegraphics[width=8cm]{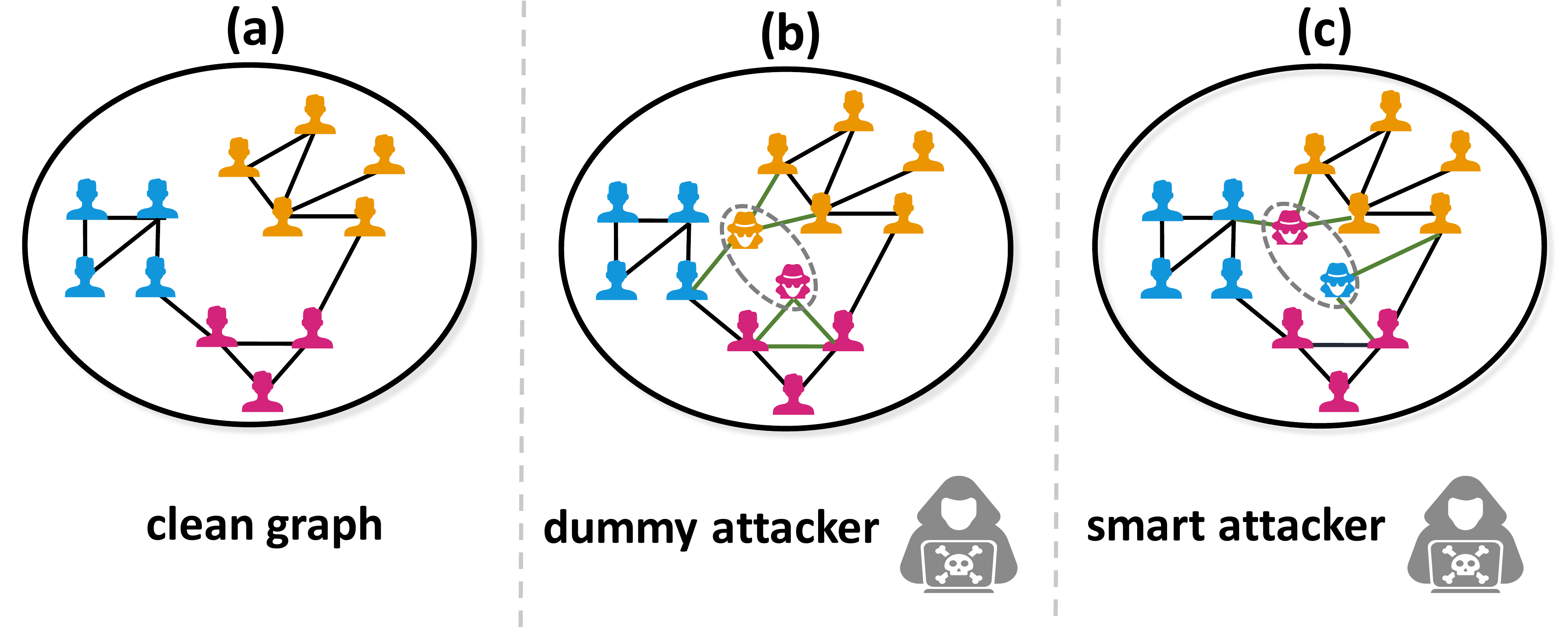}
    \vskip -1em
    \caption{(a) is the toy graph where the color of a node represents its label; (b) shows the poisoning injection attack performed by a dummy attacker; (c) shows the poisoning injection attack performed by a smart attacker. The injected nodes are circled with dashed line.}
    \label{fig:Intro}
    \vskip -2em
\end{figure}

Recent works~\cite{zugner2018adversarial, wu2019Adversarial, dai2018adversarial} have shown that even the state-of-the-art graph classifiers are susceptible to attacks which aim at adversely impacting the node classification accuracy. Because graph classifiers are trained based on node attributes and the link structure in the graph, an adversary can {\em attack} the classifier by {\em poisoning} the graph data used for training the classifier. Such an attack can be (i) node specific, as in the case of a {\em target evasion attack}~\cite{zugner2018adversarial, wu2019Adversarial} that is designed to ensure  that the node classifier is fooled into  misclassifying a specific node;  or (ii) non-target~\cite{ dai2018adversarial}, as in the case of attacks that aim to reduce the accuracy of node classification across a graph. As shown by~\cite{zugner2018adversarial, wu2019Adversarial,dai2018adversarial}, both node specific and non-target attacks can be executed  by selectively adding fake (adversarial) edges or selectively remove (genuine)  edges between the existing nodes in the graph so as to adversely impact the accuracy of the resulting graph classifiers.  However, the success of such attack strategy requires that the adversary is able to manipulate the connectivity between the nodes in the graph, e.g., Facebook, which requires breaching the security of the requisite subset of members (so as to modify their connectivity), or breaching the security of the database that stores the graph data, or manipulating the requisite members into adding or deleting their links to  other selected members. Consequently, such attack strategy is expensive for the adversary to execute without being caught.

In this paper, we introduce a novel graph {\em non-target} attack aimed at adversely impacting the accuracy of graph classifier. We describe a node injection poisoning attack procedure that provides an effective way to attack a graph  by introducing fake nodes with fake labels that link to genuine nodes so as to poison the graph data. 
Unlike previously studied graph attacks, the proposed strategy enables an attacker to boost the node misclassification rate {\em without changing the link structure between the existing nodes in the graph}. For example, in Facebook network, an attacker could simply creates fake member profiles and manipulate real members to link to the fake member profiles, so as to change the predicted labels of some of the real Facebook members. Such attack is easier and less expensive to execute compared to those that require manipulating the links between genuine nodes in the graph.

Establishing links between an injected  adversarial (fake) node to existing nodes in the original graph or to other injected adversarial nodes is a non-trivial task. As shown in Figure~\ref{fig:Intro}, both the attackers in (b) and (c) want to inject two fake nodes into the clean graph in (a). However, it is obviously presented in Figure~\ref{fig:Intro} that the "smart attacker" who carefully designs the links and label of the injected nodes could better poison the graph than the "dummy attack" who generates the links and labels at random.
We also observe that such task is naturally formulated as a {\em Markov decision process} (MDP) and reinforcement learning algorithms, e.g., Q-learning~\cite{watkins1992q} offers a natural framework for solving such problems~\cite{cai2017real, wei2017reinforcement, sutton2018reinforcement}.  However, a representation that directly encodes graph structure as states and addition and deletion of links leads to an intractable MDP. Hence, we adopt a hierarchical Q-learning network (HQN) to learn and exploit a compact yet accurate encoding of the Q function to manipulate the labels of adversarial nodes as well as their connectivity to other nodes. We propose a framework named \ourModelName  to execute the \underline{N}ode \underline{I}njection \underline{P}oisoning \underline{A}ttack.
Training the \ourModelName presents some non-trivial challenges: (i) \ourModelName has to sequentially guide fake nodes to introduce fake links to other (fake or genuine) nodes and then adversarially manipulate the labels of fake nodes; (ii) The reward function needs to be carefully designed to steer the \ourModelName to execute effective NIA.

The key contributions of the paper are as follows:
\begin{itemize}
    \item We study the novel {\em non-target} graph node injection attack problem to adversely impact the accuracy of a node classifier {\em without manipulating the link structure of the original graph}. 
    \item We propose a new framework \ourModelName, a hierarchical Q-learning  based method that can be executed by an adversary to effectively perform the poisoning attack. \ourModelName successfully addresses several non-trivial challenges presented by the resulting  reinforcement learning problem. 
    \item We present results of experiments with several real-world graphs data that show that \ourModelName outperforms the state-of-the-art non-target attacks on graph. 
\end{itemize}

The rest of the paper is organized as follows. Section 2 reviews the related work. Section 3 formally defines the  non-target node injection poisoning attack problem. Section 4 gives the details of the proposed \ourModelName. Section 5 shows empirical evaluation with discussion and section 6 presents the conclusion and future work.

\section{Related Work}
Our study falls in the general area of data poisoning attack~\cite{biggio2018wild}, which aims at attack the model by corrupting the training data. Data poisoning attacks have been extensively studied for the non graph-structured data, including supervised learning~\cite{biggio2012poisoning, mei2015Machine, li2016data}, unsupervised feature selection~\cite{xiao2015feature}, and reinforcement learning~\cite{gleave2019adversarial,jun2018adversarial,ma2018data} etc. However, little attention has been given to understanding how to poison the graph structured data.
\subsection{Adversarial Attacks on Graphs}
The previous works~\cite{szegedy2013intriguing, goodfellow2014explaining} have shown the intriguing properties of neural networks as they are "vulnerable to adversarial examples" in computer vision domain. For example, in~\cite{goodfellow2014explaining}, the authors show that some deep models are not resistant to adversarial perturbation and propose the Fast Gradient Sign Method (FGSM) to generate the adversarial image samples to attack such models. Not only in computer vision domain, recently such "intriguing properties" have also been observed in the graph mining domain. The research communities show that graph neural networks are also vulnerable to adversarial attacks. Nettack~\cite{zugner2018adversarial} is one of the first methods that perturbs the graph data to perform poisoning/training-time attack on GCN~\cite{kipf2016semi} model. RL-S2V~\cite{dai2018adversarial} adopts reinforcement learning for evasion/testing-time attack on graph data. Different from previous methods, \cite{chen2018fast} and \cite{wu2019Adversarial} focus on poison attack by gradient information. \cite{chen2018fast} attacks the graph in embedding space by iteratively modifying the connections of nodes with maximum absolute gradient. \cite{wu2019Adversarial} proposes to attack the graph structured data by use the integrated gradients approximating the gradients computed by the model and performs perturbation on data by flipping the binary value. \cite{zugner_adversarial_2019} modifies the training data and performs poisoning attacks via meta-learning. Though these graph adversarial attacks are effective, they focus on manipulating links among existing nodes in a graph, which are impractical as these nodes/individuals are not controlled by the attacker. 

Our framework is inherently different from existing work. Instead of manipulating links among existing nodes, our framework inject fake nodes to the graph (say fake accounts in Facebook), and manipulate the label and links of fake nodes to poison the graph.

\subsection{Reinforcement Learning in Graph}
Reinforcement learning(RL) has achieved significant successes in solving challenging problems such as continuous robotics control~\cite{schulman2015trust} and playing atari games~\cite{mnih2015human}. However, there has been little previous work exploring RL in graph mining domain. Graph Convolutional Policy Network (GCPN)~\cite{you2018graph} is one of the works which adopts RL in graph mining. The RL agent is trained on the chemistry aware graph environment and learns to generate molecular graph. \cite{do2019graph} is another work which defines chemical molecular reaction environment and trains the RL agent for predicting products of the chemical reaction. 
The most similar work to ours is RL-S2V~\cite{dai2018adversarial} which adopts RL for target evasion attack on graph by manipulating the links among existing nodes; while we investigate RL for non-target injection poisoning attack and manipulate labels and links of fake nodes. 
\section{Problem Definition}
In this section, we formally define the problem we target. We begin by introducing the definition of semi-supervised node classification as we aim to poison the graph for manipulating label classification of graph classifiers. Note that the proposed framework is a general framework which can also be used to poison the graph for other tasks. We leave other tasks as future work.

\begin{definition}(Semi-Supervised Node Classification) Let $G = (V,E,X)$ be an attributed graph, where  $V=\{v_1, \dots v_n\}$ denotes the node set, $E \subseteq V\times V$ means the edge set and $X$ represents the nodes features. $\mathcal{T}=\{v_{t_1},\dots, v_{t_n}\}$ is the labeled node set and $\mathcal{U}=\{v_{u_1},\dots, v_{u_n}\}$ is the unlabeled node set  with $\mathcal{T} \cup \mathcal{U}=V$. Semi-supervised node classification task aims at correctly labeling the unlabeled nodes in $\mathcal{U}$ with the graph classifier $\mathcal{C}$.
\end{definition}

In semi-supervised node classification task, the graph classifier $\mathcal{C}(G)$ which learns the mapping $ V \mapsto \tilde{L}$ aims to correctly assign the label to node $v_j \in \mathcal{U}$ with aggregating the structure and feature information. The classifier $\mathcal{C}$ is parameterized by $\theta$ and we denote the classifier as $\mathcal{C}_\theta$.  For simplicity of notations, we use $\mathcal{C}_\theta(G)_i$ as the classier prediction on $v_i$ and $\mathcal{T}_i$ as the ground truth label of $v_i$. In the training process, we aim to learn the optimal classifier $\mathcal{C}$ with the corresponding parameter $\theta_L$ defined as following:
\begin{equation}\label{eq:optimalclassifier}
\theta_L=\argmin_{\theta} \sum_{v_i\in \mathcal{T}} \mathcal{L}(\mathcal{T}_i, \mathcal{C}_\theta(G)_i)
\end{equation}

where $\mathcal{L}$ is the loss function such as cross entropy.
To attack the classifier, there are mainly two attacking settings including poisoning/training-time attack and evasion/testing-time attack. 

In poisoning attacks, the classifier $\mathcal{C}$ uses the poisoned graph for training while in evasion attack, adversarial examples are included in testing samples after $\mathcal{C}$ is trained on clean graph. In this paper, we focus on non-targeted graph poisoning attack problem where the attacker $\mathcal{A}$ poisons the graph before training time to reduce the performance of graph classifier $\mathcal{C}$ over unlabeled node set $\mathcal{U}$.

\begin{definition}(Graph Non-Targeted Poisoning Attack) Given the attributed graph $G=(V,E,X)$, the labeled node set $\mathcal{T}$, the unlabeled node set $\mathcal{U}$ and the graph classifier $\mathcal{C}$, the attacker $\mathcal{A}$ aims to modify the graph $G$ within a budget $\Delta$ to reduce the accuracy of classifier $\mathcal{C}$ on $\mathcal{U}$.
\end{definition}

As the attacking process is supposed to be unnoticeable, the number of allowed modifications of attacker $\mathcal{A}$ on $G$ is constrained by the budget $\Delta$. Based on the problem, we propose the node injection poisoning method to inject a set of adversarial nodes $V_{\mathcal{A}}$ into the node set $V$ to perform graph non-targeted poisoning attack.

\begin{definition}(Node Injection Poisoning Attack) Given the clean graph $G=(V,E,X)$, the attacker $\mathcal{A}$ injects the poisoning node set $V_{\mathcal{A}}$ with its adversarial features $X_\mathcal{A}$ and labels $\mathcal{T}_\mathcal{A}$ into the clean node set $V$. After injecting $V_{\mathcal{A}}$, the attack $\mathcal{A}$ creates adversarial edges $E_{\mathcal{A}} \subseteq  V_{\mathcal{A}} \times V_{\mathcal{A}}\cup V_{\mathcal{A}}\times V$ to poison $G$.  $G'=(V',E', X')$ is the poisoned graph where $V'=V\cup V_{\mathcal{A}}$, $E'=E\cup E_{\mathcal{A}}$, $X'=X\oplus X_\mathcal{A}$ with $\oplus$ is append operator and $\mathcal{T}'$ is the labeled set with $\mathcal{T}' = \mathcal{T} \cup \mathcal{T}_\mathcal{A} $. In the poisoning attack, the graph classifier is trained on poisoned graph $G'$. 
\end{definition}
With the above definitions and notations, the objective function for the non-targeted node injection poisoning attack is defined as:
\begin{eqnarray} 
\max_{E_\mathcal{A}, \mathcal{T}_\mathcal{A}} && \sum_{v_j\in \mathcal{U}}\mathbbm{1}(\mathcal{U}_j\neq \mathcal{C}_{\theta_L}(G')_j) \label{Eq:attack}  \\ 
s.t.  &&\theta_L=\argmin_{\theta} \sum_{v_i\in \mathcal{T'}} \mathcal{L}(\mathcal{T}'_i, \mathcal{C}_\theta(G')_i) \label{Eq:constrain1} \\
   && |E_\mathcal{A}| \leq \Delta \label{Eq:constrain2}
\end{eqnarray}
Here $\mathcal{U}_j$ represents the label of the unlabeled node $v_j$. If the attacker has the ground truth for the unlabeled data (unlabel is to end-user in this case), then $\mathcal{U}$ is ground truth label; if attacker doesn't have the access to the ground true, then $\mathcal{U}$ is predicted by graph classifier trained on clean graph.
$\mathbbm{1}(s)$ is the indicator function with $\mathbbm{1}(s)=1$ if $s$ is true and 0 otherwise. The attacker maximizes the prediction error for the unlabeled nodes in $\mathcal{U}$ as in Eq.~\eqref{Eq:attack}, subject to two constraints. The constrain~\eqref{Eq:constrain1} enforces the classifier is learned from the poisoned graph $G'$. 
and constrain~\eqref{Eq:constrain2} restricts the modifications of adversarial edges by the attacker in the budget $\Delta$

In this paper, we use the Graph Convolution Network (GCN)~\cite{kipf2016semi} as our graph classifier $\mathcal{C}$ to illustrate our framework as it is widely adopted graph neural model for node classification task. In the convolutional layer of GCN, nodes first aggregate information from its neighbor nodes followed by the non-linear transformation such as ReLU. 
The equation for a two-layer GCN is defined as:
\begin{equation}
    f(A,X)=\text{softmax}(~\hat{A}~\text{ReLU}~(\hat{A}XW^{(0)})W^{(1)})
    \label{eq:GCN}
\end{equation}
where $\hat{A}=\hat{D}^{-\frac{1}{2}}\tilde{A}\hat{D}^{-\frac{1}{2}}$ denotes the normalized adjacency matrix, $\tilde{A} = A+I_N$ denotes adding the identity matrix $I_N$ to the adjacent matrix $A$. $\hat{D}$ is the diagonal matrix with on-diagonal element as $\hat{D}_{ii} = \sum_j\tilde{A}_{ij}$. $W^{(0)}$ and $W^{(1)}$ are the weights of first and second layer of GCN, respectively. $\text{ReLU(0, a)} = \max(0, a)$ is adopted. The loss function $\mathcal{L}$ in GCN is cross entropy.

\section{Proposed Framework}
\begin{figure*}[ht]
    \centering
    \includegraphics[width=14cm]{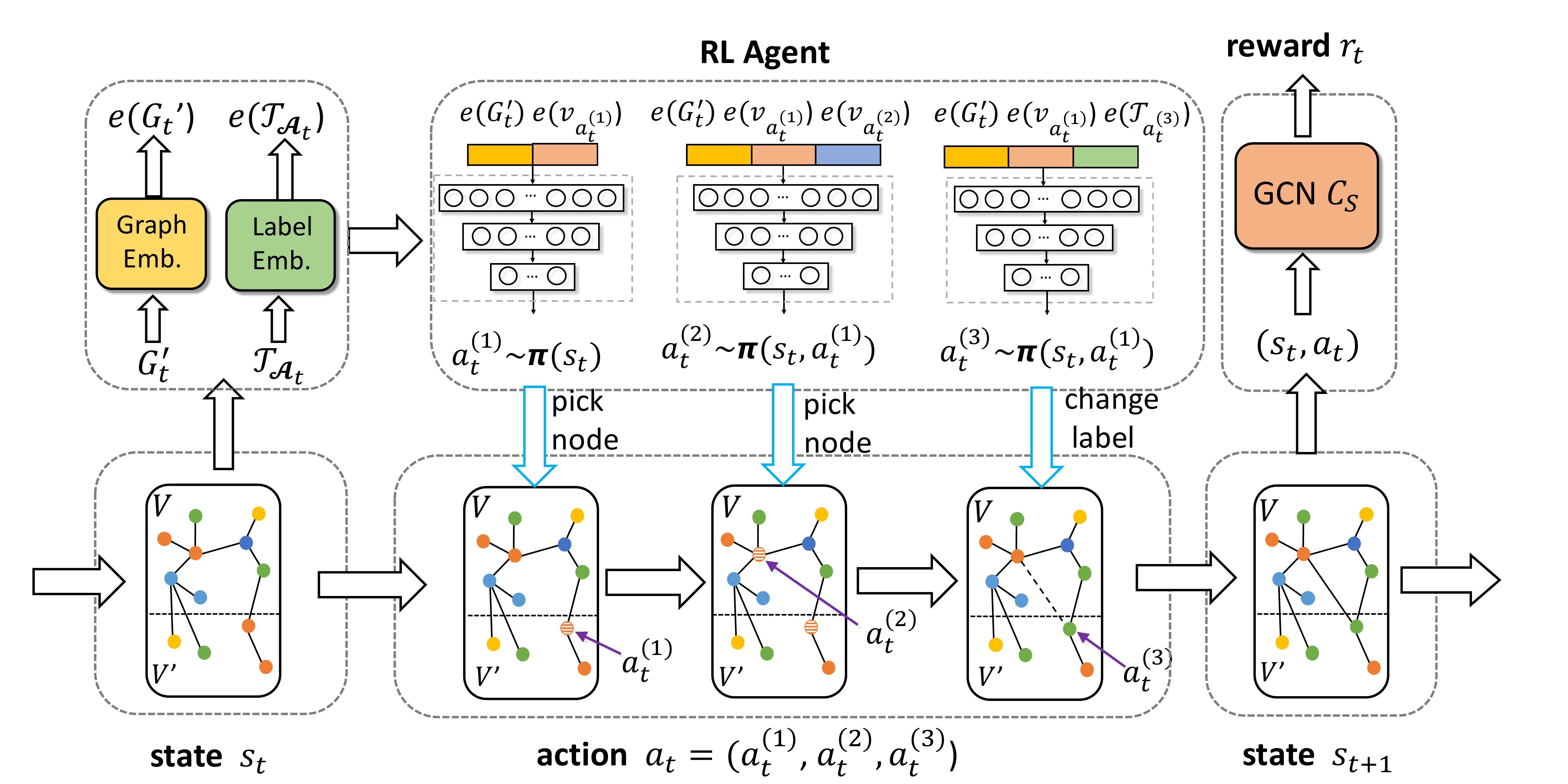}
    \vskip -1em
    \caption{An overview of the Proposed Framework \ourModelName for Node Injection Attack on Graphs}
    \label{fig:framework}
    \vskip -1em
\end{figure*}

To perform the non-target node injecting poisoning attack, we propose to solve the optimization problem in Eq.(\ref{Eq:attack}) via deep reinforcement learning.  
Compared with directly optimizing the adjacency matrix with traditional matrix optimization techniques, the advantages of adopting deep reinforcement learning are two folds: (\textbf{i}) Adding edges and changing labels of fake nodes are naturally sequential decision making process. Thus, deep reinforcement learning is a good fit for the problem; (\textbf{ii}) The underlying structures of graphs are usually highly nonlinear~\cite{wang2016structural}, which adds the non-linearity to the decision making process. The deep non-linear neural networks of the Q network could better capture the graph non-linearity and learn the semantic meaning of the graph for making better decisions.

An illustration of the proposed framework is shown in Figure~\ref{fig:framework}. The key idea of our proposed framework is to train the deep reinforcement learning agent which could iteratively perform actions to poison the graph. The actions includes adding adversarial edges and modifying the labels of injected nodes. More specifically, the agent needs to firstly pick one node from injected nodes set $V_\mathcal{A}$ and select another node from poisoned node set $V'$ to add the adversarial edge, and modify the label of the injected nodes to attack the classifier $\mathcal{C}$. We design reinforcement learning environment and reward according to the optimization function to achieve this.

Next, we describe the details of the proposed method and present the RL environment design, the deep Q

\subsection{Attacking Environment}
We model the proposed poisoning attack procedure as a Finite Horizon Markov Decision Process $(S,A,P,R,\gamma)$. 
The definition of the MDP contains state space $S$, action set $A$, transition probability $P$, reward $R$, discount factor $\gamma$. 

\subsubsection{State}
The state $s_t$ contains the intermediate poisoned graph $G_t'$ and labels information $\mathcal{T}_{\mathcal{A}_t}$ of the injected nodes at the time $t$. To capture the highly non-linear information and the non-Euclidean structure of the poisoned graph $G_t'$, we embed $G_t'$ as $e(G_t')$ with aggregating the graph structure information via designed graph neural networks. 
$e(\mathcal{T}_{\mathcal{A}_t})$ encodes the adversarial label information $L_{\mathcal{A}_t}$ with neural networks. The details of the state representation is described in following subsection. 

Since in the injection poisoning environment, the node set $V'$ remains identical thus the DRL agent performs poisoning actually  on the edge set $E'_t$.

\subsubsection{Action}
In the poisoning attack environment, the agent is allowed to (1) add the adversarial edges within the injected nodes $V_{\mathcal{A}}$ or between the injected nodes and the clean nodes; (2) modify the adversarial labels of the injected nodes. However, directly adding one adversarial edge has $O(|V_{\mathcal{A}}|^2+|V_{\mathcal{A}}|*|V|)$ possible choices and modifying the adversarial label of one injected node requires $O(|L|)$  space where $|L|$ is the number of label categories.
Thus, performing one action that contains both adding an adversarial edge and changing the label of a node has search space as $O(|V_{\mathcal{A}}|*|V'|*|L|)$, which is extremely expensive especially in large graphs. Thus, we adopt hierarchical action to decompose such action and reduce the action space to enable efficient exploration inspired by previous work~\cite{dai2018adversarial}.

As shown in Figure~\ref{fig:framework}, in \ourModelName, the agent first performs an action $a^{(1)}$ to select one injected node from $V_\mathcal{A}$. The agent then picks another node from the whole node set $V'$ via action $a^{(2)}$. The agent connects these two selected nodes to forge the adversarial edge. Finally, the agent modifies the label of the selected fake node through action $a_3$. By such hierarchical action $a = (a^{(1)}, a^{(2)}, a^{(3)})$, the action space is reduced from  $O(|V_{\mathcal{A}}|*|V'|*|L|)$ to $O(|V_{\mathcal{A}}|+|V'|+|L| )$. 
With the hierarchy action  $a = (a^{(1)}$, $a^{(2)}$, $a^{(3)})$, the trajectory of the proposed MDP is $(s_0,a^{(1)}_0, a^{(2)}_0, a^{(3)}_0, r_0, $ $s_1,\dots,s_{T-1},a^{(1)}_{T-1},a^{(2)}_{T-1}, a^{(3)}_{T-1}, r_{T-1}, s_{T})$.

\subsubsection{Policy network}
As both of previous work~\cite{dai2018adversarial} and our our preliminary experiments show that Q-learning
works more stable than other policy optimization methods such as Advantage Actor Critic, we focus on modeling policy network with Q-learning.
Q-learning is an off-policy optimization which fits the Bellman optimality equation as:
\begin{equation}
    Q^*(s_t,a_t) =  r(s_t,a_t)+\gamma \max_{a'_t}Q^*(s_{t+1}, a')
    \label{eq:qnetwork}
\end{equation}
The greedy policy to select the action $a_t$ with respect to $Q^*$ is: 
\begin{equation}
    a_t = \pi(s_t)=\argmax_{a}Q^*(s_t,a)
    \label{eq:policy}
\end{equation}

As we explain in the above subsection that performing one poisoning action requires searching in $O(|V_{\mathcal{A}}|*|V'|*|L|)$ space and we perform hierarchical actions other than one action, we cannot directly follow the policy network in Eq.(\ref{eq:qnetwork}) and Eq.(\ref{eq:policy}).
Here, we adopt hierarchical Q function for the actions and we propose the hierarchical framework which integrates three DQNs. The details of the proposed DQNs are presented in following section.

\subsubsection{Reward}
As the RL agent is trained to enforce the misclassification of the graph classifier $\mathcal{C}$, we need to design the reward accordingly to guide the agent.
The reasons why we need to design novel reward function other than using the widely adopted binary sparse rewards are two folds: (1) as our trajectory in the attacking environment is usually long, we need the intermediate rewards which give feedback to the RL agent on how to improve its performance on each state; (2) different from the target attack that we know whether the attack on one targeted node is success or not, we perform the non-target attack over graph thus accuracy is not binary
The reward of the current state and actions for the agent is designed according to the poisoning objective function shown in Eq.~\eqref{Eq:attack}. For each state  $s_t$, we define the attack success rate $\mathcal{A}_t$ as:

\begin{equation}\label{eq:reward overview}
 \mathcal{A}_t =  \sum_{v_j\in \mathcal{D}}\mathbbm{1}(\mathcal{D}_j\neq \mathcal{C}_{\theta_S}(G'_t)_j)/|\mathcal{V}|
\end{equation}
\begin{equation}
    \theta_S=\argmin_{\theta} \sum_{v_i\in \mathcal{T'}} \mathcal{L}(\mathcal{T}'_i, \mathcal{C}_\theta(G')_i) 
\end{equation}
Here $\mathcal{D}$ is the validation set used to compute the reward.
Note that the $\mathcal{C}_{\theta_S}$ is not the graph classifier $\mathcal{C}$ that evaluates the final classification accuracy. It represents the simulated graph classifier designed by attacker to acquire the state and actions reward.
However, directly using the success rate $\mathcal{A}_t$ as the reward would increase the instability of training process since the accuracy might not differ a lot for two consecutive state. In this case, we design the guiding binary reward $r_t$ to be one if the action $a_t = (a^{(1)}_t, a^{(2)}_t,  a^{(3)}_t)$ could reduce the accuracy of attacker's simulated graph classifier $\mathcal{C}_{\theta_S}$, and to be negative one vice versa. The design the guiding reward $r_t$ is defined as follows:
\begin{equation}
   \label{eq:reward design}
r_t(s_t, a^{(1)}_t, a^{(2)}_t,  a^{(3)}_t) = \left\{  
\begin{aligned}
& \; 1; \quad \text{if}\;\mathcal{A}_{t+1} > \mathcal{A}_{t} \\
&{-1}; \quad \text{otherwise}.\\
\end{aligned}
\right.  
\end{equation}
Our preliminary experimental results show that such guiding reward is effective in our case.
\subsubsection{Terminal}
In the poisoning attacking problem, the number of allowed adding adversarial edges is constrained by the budget $\Delta$ for the unnoticeable consideration. So in the poisoning reinforcement learning environment, once the agent adds $T=\Delta$ edges, it stops taking actions. In terminal state $s_T$,  the poisoned graph $G'$ contains $T$ more adversarial edges compared to the clean graph $G$.

\subsection{State Representation}
As mentioned above, the state $s_t$ contains the poisoned graph $G_t'$ and injected nodes labels $\mathcal{T}_{\mathcal{A}_t}$ at time $t$. 
To represent the non-Euclidean structure of the poisoned graph $G_t'$ with vector $e(G'_t)$, the latent embedding $e(v_i)$ of the each node $v_i$ in $G_t'$ is firstly learned by struct2vec~\cite{dai2016discriminative} using the discriminative information. Then the state vector representation $e(G_t')$ is obtained by aggregating the embedding of nodes as:
\begin{equation}\label{eq:graph embedding}
   e(G'_t) = \sum_{v_i \in V'}e(v_i)/|V'|
\end{equation}
To represent the label of the injected nodes, we use the two layer neural networks to encode the $\mathcal{T}_{\mathcal{A}_t}$ as $e(\mathcal{T}_{\mathcal{A}_t})$.
Note that for the notation compact and consistency consideration, $e(s)$ represents embedding of the state, and $e(v_a)$ and $e(\mathcal{T}_a)$ are the embeddings of the node selected by action $a$ and label selected by action $a$ respectively in the following paper.

\subsection{Hierarchical Q Network}
In Q learning process,  given the state $s_t$ and action $a_t$, the action-value function $Q(s_t,a_t)$ is supposed to give the scores of current state and selected actions to guide the RL agent. However, as the action $a$ is decomposed into three hierarchical actions $\{a^{(1)}, a^{(2)}, a^{(3)}\}$ for the efficiency searching consideration, it would be hard to directly design the $Q(s_t, a^{(1)}_t, a^{(2)}_t, a^{(3)}_t)$ and apply one policy network to select hierarchical actions.

To overcome this problem, we adopt hierarchical deep Q networks $Q=\{Q^{(1)}, Q^{(2)}, Q^{(3)}\}$ which integrates three DQNs  to model the Q values over the actions.
Figure~(\ref{fig:framework}) illustrates the framework of selection action $a_t= \{a^{(1)}_t, a^{(2)}_t, a^{(3)}_t)\}$ at time $t$.  
The first DQN $Q^{(1)}$ guides the policy to select a node from injected node set  $V_\mathcal{A}$; based on $a_t^{(1)}$, the second DQN  $Q^{(2)}$ learns the policy to select a second node from the node set $V'$, which completes an edge injection by connecting the two nodes. The third DQN $Q^{(3)}$ learns the policy to modify the label of the first selected injected node.

The agent firstly selects one node from the injected node set $V_{\mathcal{A}}$ and calculate the $Q$ value based on the action-value function $Q^{(1)}$ as: 
\begin{equation}\label{eq:dqn-1}
    Q^{(1)}(s_t,a^{(1)}_t;\theta^{(1)})=W^{(1)}_{1}\sigma(W^{(1)}_{2}[ e(s_t) \mathbin\Vert e(v_{a^{(1)}_t})]) 
\end{equation}
where $\theta^{(1)} = \{ W_1^{(1)}, W^{(1)}_{2}\}$ represents the trainable weights of the first DQN and $\mathbin\Vert$ is the concatenation operation. The action-value function $Q^{(1)}$ estimates the Q value given the state and action. The greedy policy to select the action $a_t^{(1)}$ based on optimal action-value function $Q^{(1)*}$ in eq.(\ref{eq:dqn-1}) is defined as follows:

\begin{equation}\label{eq:policy-1}
    a^{(1)}_t = \pi(s_t)=\argmax_{a\in V_\mathcal{A}}Q^{(1)}(s_t,a;\theta^{(1)}); \;  
\end{equation}

With the first action $a^{(1)}_t$ selected, the agent picks the second action $a^{(2)}_t$ hierarchically based on  $Q^{(2)}$ as:
\begin{equation}\label{eq:dqn-2}
   Q^{(2)}(s_t,a^{(1)}_t, a^{(2)}_t; \theta^{(2)})=W^{(2)}_{1}\sigma(W^{(2)}_{2}[ e(s_t) \mathbin\Vert e(v_{a^{(1)}_t}) \mathbin\Vert e(v_{a^{(2)}_t})])
\end{equation}
where $\theta^{(2)}=\{W^{(2)}_{1}, W^{(2)}_{2}\}$ is the trainable weights. The action value function $Q^{(2)}$ scores the state, and the action $a^{(1)}_t$ and $a^{(2)}_t$. The greedy policy to make the second action $a_t^{(2)}$ with the optimal $Q^{(2)*}$ in eq.(\ref{eq:dqn-2}) is defined as follows:
\begin{equation}\label{eq:policy-2}
a^{(2)}_t = \pi(s_t, a^{(1)}_t)=\argmax_{a\in V'}Q^{(2)}(s_t,a^{(1)}_t, a;\theta^{(2)}); \; 
\end{equation}

Note that the agent only modifies the label of the selected injected node $a_t^{(1)}$, the action-value function for the third action is not related to the action $a_t^{(2)}$. The action-value function $Q^{(3)}$ is defined as follows: 
\begin{equation}\label{eq:dqn-3}
    Q^{(3)}(s_t,a^{(1)}_t,a^{(3)}_t; \theta^{(3)})=W^{(3)}_{1}\sigma(W^{(3)}_{2}[ e(s_t)\mathbin\Vert e(v_{a^{(1)}_t})\mathbin\Vert  e(\mathcal{T}_{a^{(3)}_t})]) 
\end{equation}
In Eq.(\ref{eq:dqn-3}), $\theta^{(3)}=\{W_1^{(3)}, W_2^{(3)}\}$ represents the trainable weights in $Q^{(3)}$. The action value function $Q^{(3)}$ models the score of changing the label of the injected node $a_t^{(1)}$.
The greedy policy to such action is defined as follows:
\begin{equation}\label{eq:policy-3}
a^{(3)}_t = \pi(s_t,a^{(1)}_t)=\argmax_{a \in L_\mathcal{A}}Q^{(3)}(s_t,a^{(1)}_t, a;\theta^{(3)}); \;
\end{equation}

\subsection{Training Algorithm}
To train the proposed hierarchy DQNs $Q=\{Q^{(1)}, Q^{(2)}, Q^{(3)}\}$ and the graph embedding method structure2vec, we use the experience replay technique with memory buffer $\mathcal{M}$. The high level idea is simulating the selection process to generate training data, which are stored in memory buffer, during the reinforcement learning training process.
During training, the experience $(s, a, s')$ where $a = \{a^{(1)}, a^{(2)},a^{(3)}\}$ is drawn uniformly at random from the stored memory buffer $\mathcal{M}$. The Q-learning loss function is similar to \cite{mnih2015human} as:
\begin{equation}\label{eq:loss}
\mathbb{E}_{(s,a,s')\sim \mathcal{M}}[(r+\gamma \max_{a'} \hat{Q}(s',a'|\theta^{-}) - Q(s,a|\theta))^2]
\end{equation}
where $\hat{Q}$ represents target action-value function and its parameters $\theta^{-}$ are updated with $\theta$ every C steps.
To improve the stability of the algorithm, we clip the error term between $-1$ and $+1$. The agent adopts $\epsilon$-greedy policy that select a random action with probability $\epsilon$. The overall training framework is summarized in Algorithm \ref{algo:RL}. 

\begin{algorithm}[ht]
	\caption{The training algorithm of framework \ourModelName}
    \label{algo:RL}
    \KwIn{clean graph $G(V,E, X)$, labeled node set $\mathcal{T}$, budget $\Delta$, number of injected nodes $|V_\mathcal{A}|$, training iteration $K$}  
    \KwOut{$G'(V',E',X')$ and $L_\mathcal{A}$}
	Initialize action-value function Q with random parameters $\theta$ \;
	Set target function $\hat{Q}$ with parameters $\theta^{-}=\theta$\;
	Initialize replay memory buffer $\mathcal{M}$\;
	Randomly assign Adversarial label $L_\mathcal{A}$\;
	\While{episode $\leq K$ }{
	\While{$t\leq \Delta$}{
    	    Select $a_t^{(1)}$ based on  Eq.(\ref{eq:dqn-1})\;
    	    Select $a_t^{(2)}$ and $a_t^{(3)}$ based on  Eq.(\ref{eq:dqn-2}) and Eq.(\ref{eq:dqn-2})\;
    	    Compute $r_t$ according to Eq.(\ref{eq:reward overview}) and Eq.(\ref{eq:reward design})\;
    	    Set $s_{t+1}=\{s_t, a_t^{(1)}, a_t^{(2)}, a_t^{(3)}\}$\;
    	    $E_\mathcal{A} \leftarrow E_\mathcal{A} \cup (a_t^{(1)}, a_t^{(2)})$, $L_{\mathcal{A}a_t^{(1)}}\leftarrow a_t^{(3)}$  \;
    	    Store $\{s_t, a_t^{(1)}, a_t^{(2)}, a_t^{(3)}, r_t,s_{t+1}\}$ in memory $\mathcal{M}$\;
    	    Sample minibatch transition randomly from $\mathcal{M}$\;
    	    Update parameter according to Eq.(\ref{eq:loss})\;
    	   Every C steps $\theta^{-}=\theta$;
	    }
	    }
\end{algorithm}
In the proposed model, we use two layer multi-layer perceptrons to implement all the trainable parameters $\theta$ in action-value functions $Q^{(1)}$, $Q^{(2)}$, $Q^{(3)}$ and structure2vec. Actually, more complex deep neural networks could replace the models outlined here. We leave exploring feasible deep neural networks as a future direction.
\section{Experiments}
In this section, we introduce the experiment setting including baseline datasets and comparing poisoning attack methods. Moreover, we conduct experiments and present results to answer the following research questions: (\textbf{RQ1}) Can the \ourModelName effectively poison the graph data via node injection? (\textbf{RQ2}) Whether the poisoned graph remains the key statistics after the poison attack?  (\textbf{RQ3}) How the proposed framework performances under different scenarios? Next, we first introduce the experimental settings followed by experimental results to answer the three questions.

\subsection{Experiment Setup}
\subsubsection{Datasets}
We conduct experiments on three widely used  benchmark datasets for node classification, which include CORA-ML~\cite{mccallum2000automating,bojchevski2018deep}, CITESEER~\cite{giles1998citeseer} and DBLP~\cite{pan2016tri}. 
Following~\cite{zugner_adversarial_2019}, we only consider the largest connected component (LCC) of each graph data. The statistics for the datasets are summarized in
Table \ref{table:statistics}.  For each dataset, we randomly split the nodes into (20\%) labeled nodes for training procedure and (80\%) unlabeled nodes as test set to evaluate the model. The labeled nodes are further equally split into training and
validation sets. We perform the random split five times and report averaged results. 
\begin{table}[ht]
\caption{Statistics of benchmark datasets}
\vskip -1.5em
\small
\begin{tabular}{cccccc}
\toprule
Datasets & $N_{LCC}$ & $E_{LCC}$ &|L|\\ \midrule
CITESEER & 2,110   &3,757    &6      \\
CORA-ML  & 2,810    &7,981 &7     \\ 
PUBMED & 19,717&44,324 &3\\
\bottomrule
\end{tabular}
\label{table:statistics}
\end{table}

\subsubsection{Baseline Methods}
Though there are several adversarial attack algorithms on graphs such as Nettack~\cite{zugner2018adversarial}  and RL-S2v~\cite{dai2018adversarial}, most of them are developed for manipulating links among existing nodes in graph, which cannot be easily modified in our attacking setting for node injection attack. Thus, we don't compare with them.  Since node injection attack on graphs is a novel task, there are very few baselines we can compare with. We select  following four baselines, with two from classical graph generation models, one by applying the technique of fast gradient attack and a variant of \ourModelName.
\begin{itemize}[leftmargin=*]
    \item Random Attack: The attacker $\mathcal{A}$ first adds adversarial edges between the injected nodes according to Erdos-Renyi model~\cite{erdHos1960evolution} $G(V_\mathcal{A},p)$, where the probability $p=\frac{2|E|}{|V|^2}$ is the average degree of the clean graph $G(V, E)$ to make sure the density of the injected graph $G_A$ is similar to the clean graph. The attacker then randomly add adversarial edges connecting the injected graph $G_A$ and clean graph $G$ until the budget $\Delta$ is used ups.
    \item Preferential attack~\cite{barabasi1999emergence}: The attacker $\mathcal{A}$ iteratively adds the adversarial edges according to preferential attachment mechanism. The probability of connecting the injected node $v_i \in  V_\mathcal{A}$ to the other node $v_j \in |V\cup V_\mathcal{A}|$ is proportional to the node degrees. The number of adversarial edges is constrained by the budget $\Delta$.
    \item Fast Gradient Attack(FGA)~\cite{chen2018fast}: Gradient based methods are designed to attack the graph data by perturbing the gradients. In FGA, the attacker $\mathcal{A}$ removes/adds the adversarial edges guided by edge gradient. 
    \item \ourModelName-w/o: This is a variant of the proposed framework \ourModelName where we don't optimize w.r.t the label of fake nodes, i.e., the labels of the fake nodes are randomly assigned.   
\end{itemize}

\begin{table*}[h]
\caption{classification results after attack}
\vskip -1.5em
\small
\begin{tabular}{c | c | cccccc}
\toprule
Dataset & Methods &$r=0.01$ & $r=0.02$ &$r=0.05$ &$r=0.10$\\
\midrule
& Random  &0.7582 $\pm$ 0.0082 &0.7532 $\pm$ 0.0130 &0.7447 $\pm$ 0.0033 &0.7147 $\pm$ 0.0122 \\
CITESEER & Preferrential  &0.7578 $\pm$ 0.0060 &0.7232 $\pm$ 0.0679 &0.7156 $\pm$ 0.0344 &0.6814 $\pm$ 0.0131  \\
&FGA &0.7129 $\pm$ 0.0159 &0.7117 $\pm$ 0.0052 &0.7103 $\pm$ 0.0214 &0.6688 $\pm$ 0.0075 \\
& \ourModelName-wo(ours)  &0.7190 $\pm$ 0.0209  &0.6914 $\pm$ 0.0227 &0.6778 $\pm$ 0.0162 &0.6301 $\pm$ 0.0182 
\\ 
&\ourModelName(ours)  &\textbf{0.7010} $\pm$ 0.0123 &\textbf{0.6812} $\pm$ 0.0313 &\textbf{0.6626} $\pm$ 0.0276 &\textbf{0.6202} $\pm$ 0.0263 \\
\midrule
 & Random  &0.8401 $\pm$ 0.0226 &0.8356 $\pm$ 0.0078 &0.8203 $\pm$ 0.0091 &0.7564 $\pm$ 0.0192 \\
CORA-ML & Preferrential   &0.8272 $\pm$ 0.0486 &0.8380 $\pm$ 0.0086 &0.8038 $\pm$ 0.0129 &0.7738 $\pm$ 0.0151 \\
&FGA &0.8205 $\pm$ 0.0044 &0.8146 $\pm$ 0.0041 &0.7945 $\pm$ 0.0117 &0.7623 $\pm$ 0.0079 \\
& \ourModelName-w/o (ours) &0.8042 $\pm$ 0.0190 &0.7948 $\pm$ 0.0197 &0.7631 $\pm$ 0.0412   &0.7206 $\pm$ 0.0381  \\
&\ourModelName(ours)   &\textbf{0.7902} $\pm$ 0.0219 &\textbf{0.7842} $\pm$ 0.0193  &\textbf{0.7461} $\pm$ 0.0276  &\textbf{0.6981} $\pm$ 0.0314 \\
\midrule
 & Random  &0.8491 $\pm$ 0.0030 &0.8388 $\pm$ 0.0035 &0.8145 $\pm$ 0.0076 &0.7702 $\pm$ 0.0126  \\
PUMBED & Preferrential  &0.8487 $\pm$ 0.0024 &0.8445 $\pm$ 0.0035 &0.8133 $\pm$ 0.0099 &0.7621 $\pm$ 0.0096 \\
&FGA &0.8420 $\pm$ 0.0182 &0.8312 $\pm$ 0.0148 &0.8100 $\pm$ 0.0217 &0.7549 $\pm$ 0.0091\\
& \ourModelName-w/o(ours)  &0.8412 $\pm$ 0.0301 &0.8164 $\pm$ 0.0209 &0.7714 $\pm$ 0.0195  &0.7042 $\pm$ 0.0810 \\
&\ourModelName (ours) &\textbf{0.8242} $\pm$ 0.0140 &\textbf{0.8096} $\pm$ 0.0155 &\textbf{0.7646} $\pm$ 0.0065  &\textbf{0.6901} $\pm$ 0.0203 \\
\bottomrule
\end{tabular}
\label{table:result}
\end{table*}

The Fast Gradient Attack(FGA)~\cite{chen2018fast} is not directly applicable in injection poisoning setting, since the injected nodes are isolated at the beginning and would be filtered out by graph classifier. Here we modify the FGA for fair comparison. The FGA method is performed on the graph poisoned by preferential attack. After calculating the gradient $\nabla_{ij} L_{GCN}$ with $v_i \in V_\mathcal{A}$ and $v_j \in V'$, the attack $\mathcal{A}$ adding/remove the adversarial edges between $(v_i, v_j )$  according to the largest positive/negative gradient. The attack only add and remove one feasible adversarial edge are each iteration so that the number of the adversarial edges is still constrained by budget $\Delta$. The attacker is allowed to perform 20*$\Delta$ times modifications in total  suggested by~\cite{chen2018fast}.

\subsection{Attack Performance Comparison}
To answer \textbf{RQ1}, we evaluate how the node classification accuracy degrades on the poisoned graph compared with the performance on the clean graph. The larger decrease the performance is on the poisoned graph, the more effective the attack is.

\vskip 0.5em
\noindent{}\textbf{Node Classification on Clean Graph} As the Nettack~\cite{zugner2018adversarial} points out that ``poisoning attacks are in general harder and match better the transductive learning scenario'', we follow the same poisoning transductive setting in this paper. The parameters of GCN are trained according to Eq.~\eqref{eq:optimalclassifier}. We report the averaged node classification accuracy over five runs in Table.~\ref{table:clean} to present the GCN node classification accuracy on clean graph. 
Note that if the poisoning nodes are injected with the budget $\Delta=0$, such isolated nodes would be filtered out by GCN and the classification results remain the same as in Table.~\ref{table:clean}.

\begin{table}[h]
\caption{Node classification results on clean graph}
\vskip -1.5em
\small
\begin{tabular}{c |ccccccc}
\toprule
Dataset  &CITESEER &CORA-ML &Pubmed\\ 
\midrule
Clean data  &0.7730 $\pm$ 0.0059 &0.8538 $\pm$ 0.0038 &0.8555 $\pm$ 0.0010\\
\bottomrule
\end{tabular}
\label{table:clean}
\end{table}

\vskip -1em
\noindent{}\textbf{Node Classification on Poisoned Graph}

In poisoning attacking process, the attacking budget $\Delta$ which controls the number of added adversarial edges is one important factor. On the one hand, if the budget is limited, eg., $\Delta < |V_\mathcal{A}|$, then at least $|V_\mathcal{A}|-\Delta$ injected nodes are isolated. Clearly, isolated nodes have no effect on the label prediction as they are not really injected into the environment. On the other hand, if the budget is large, the density of the injected graph is different from the clean graph and such injected nodes might be detected by the defense methods.  Here, to make the poisoned graph has the similar density with the clean graph and simulates the real world poisoning attacking scenario, we set $\Delta=r*|V|*{deg}(V)$ where $r$ is the injected nodes ratio compared to the clean graph and $deg(V)$ is the average degree of the clean graph $G$. The injected nodes number is $|V_\mathcal{A}|=r*|V|$. We will evaluate how effective the attack is when the injected nodes can have different number of degrees in Section~\ref{sec:degree_effect}. 
To have comprehensive comparisons of the methods, we vary $r$ as $r = \{0.01,0.02,0.05,0.10\}$. We don't set $r > 0.10$ since we believe that too much injected nodes could be easily noticed in real-world scenarios. 
For the same unnoticeable consideration, the feature of the injected nodes is designed to be similar to that of the clean node features. For each injected node, we calculate the mean of the features as $\bar X$ and apply the Gaussian noise $\mathcal{N}(0,1)$ on the averaged features $\bar X$. The features of the injected node are similar to the features in clean graph. We leave the generation of node features as future work. As the other baselines method could not modifies the adversarial labels of the injected nodes, we also provide the variant model \ourModelName-w/o  which doesn't manipulate the adversarial labels for fair comparison. The adversarial labels are randomly generated within $|L|$ for the baseline methods. In both \ourModelName and \ourModelName-w/o, we set the discount factor $\gamma=0.9$ and the injected nodes $V_\mathcal{A}$ are only appear in training phase in all of the methods. 

The averaged results with standard deviation for all methods are reported in Table~\ref{table:result}. From Table~\ref{table:clean} and \ref{table:result}, we could observe that (1) In all attacking methods, more injected nodes could better reduce the node classification accuracy, which satisfy our expectation.  (2) Compared with Random and Preferential attack, FGA is relatively more effective in attacking the graph, though the performance gap is marginal. This is because random attack and preferential attack don't learn information from the clean graph and just insert fake nodes following predefined rule. Thus, both of the methods are not as effective as FGA which tries to inject nodes through a way to decrease the performance. (3) The proposed framework outperforms the other methods. In particular, both FGA and NIPA are optimization based approach while NIPA significantly outperforms FGA, which implies the effectiveness of the proposed framework by designing hierarchical deep reinformcent learning to solve the decision making optimization problem. 
(4) NIPA out performances NIPA-w/o, which shows the necessity of optimizing w.r.t to labels for node injection attack.

\begin{table*}[ht]
\caption{Statistics of the clean graph $(r=0.00)$ and the graphs poisoned by \ourModelName averaged over 5 runs.}
\vskip -1em
\small
\begin{tabular}{cccccccccccccccccc}
\toprule
Dataset & $r$ &Gini Coefficient &Characteristic Path Length &Distribution Entropy &Power Law Exp.  &Triangle Count \\
\midrule
&0.00 &0.3966 $\pm$ 0.0000 &6.3110 $\pm$ 0.0000   &0.9559 $\pm$ 0.0000 &1.8853 $\pm$ 0.0000 &1558.0 $\pm$ 0.0 \\
&0.01 &0.4040 $\pm$ 0.0007 &6.0576 $\pm$ 0.1616  &0.9549 $\pm$ 0.0004 &1.8684 $\pm$ 0.0016 &1566.2 $\pm$ 7.4  \\
CORA &0.02 &0.4075 $\pm$ 0.0002 &6.1847 $\pm$ 0.1085  &0.9539 $\pm$ 0.0002 &1.8646 $\pm$ 0.0006 &1592.0 $\pm$ 17.4 \\
& 0.05 &0.4267 $\pm$ 0.0014 &5.8165 $\pm$ 0.1018  &0.9458 $\pm$ 0.0009 &1.8429 $\pm$ 0.0027 &1603.8 $\pm$ 12.8 \\
& 0.10 &0.4625 $\pm$ 0.0005 &6.1397 $\pm$ 0.0080  &0.9261 $\pm$ 0.0007 &1.8399 $\pm$ 0.0017 &1612.4 $\pm$ 22.2\\
\midrule
&0.00 &0.4265 $\pm$ 0.0000 &9.3105 $\pm$ 0.0000 &0.9542 $\pm$ 0.0000 &2.0584 $\pm$ 0.0000 &1083.0 $\pm$ 0.0 \\
& 0.01 &0.4270 $\pm$ 0.0012 &8.3825 $\pm$ 0.3554 &0.9543 $\pm$ 0.0001 &2.0296 $\pm$ 0.0024   &1091.2 $\pm$ 6.6\\
CITESEER& 0.02 &0.4346 $\pm$ 0.0007 &8.3988 $\pm$ 0.2485 &0.9529 $\pm$ 0.0005 &2.0161 $\pm$ 0.0007  &1149.8 $\pm$ 32.4 \\
 &0.05 &0.4581 $\pm$ 0.0026 &8.0907 $\pm$ 0.7710 &0.9426 $\pm$ 0.0009 &1.9869 $\pm$ 0.0073 &1174.2 $\pm$ 42.8  \\
 &0.10 &0.4866 $\pm$ 0.0025 &7.3692 $\pm$ 0.6818 &0.9279 $\pm$ 0.0012 &1.9407 $\pm$ 0.0088 &1213.6 $\pm$ 61.8\\
\midrule
& 0.00 &0.6037 $\pm$ 0.0000 &6.3369 $\pm$ 0.0000 &0.9268 $\pm$ 0.0000 &2.1759 $\pm$ 0.0000  &12520.0 $\pm$ 0.0 \\
& 0.01 &0.6076 $\pm$ 0.0005 &6.3303 $\pm$ 0.0065 &0.9253 $\pm$ 0.0004 &2.1562 $\pm$ 0.0013&12570.8 $\pm$ 29.2  \\
PUBMED &0.02 &0.6130 $\pm$ 0.0006 &6.3184 $\pm$ 0.0046  &0.9213 $\pm$ 0.0004 &2.1417 $\pm$ 0.0009  &13783.4 $\pm$ 101.8  \\
&0.05  &0.6037 $\pm$ 0.0000 &6.3371 $\pm$ 0.0007  &0.9268 $\pm$ 0.0000 &2.1759 $\pm$ 0.0001  &14206.6 $\pm$ 152.8 \\
&0.10 &0.6035 $\pm$ 0.0003 &6.2417 $\pm$ 0.1911   &0.9263 $\pm$ 0.0010 &2.1686 $\pm$ 0.0141  &14912.0 $\pm$ 306.8 \\
\bottomrule
\label{table:poisoned statistic}
\end{tabular}
\vskip -1em
\end{table*}

\subsection{Key Statistics of the Poisoned Graphs}
To answer \textbf{RQ2}, we analyze some key statistics of the poisoned graphs, which helps us to understand the attacking behaviors. One desired property of the poisoning graph is that the poisoned graph has similar graph statistics to the clean graph. We use the same graph statistics as that used in~\cite{bojchevski2018netgan} to measure the poisoned graphs for the three datasets. The results are reported in Table~\ref{table:poisoned statistic}. It could be concluded from the graph statistics that (1) Poisoned graph has very similar graph distribution to the clean graph. For example, the similar exponent of the power law distribution in graph indicates that the poisoned graph and the clean graph shares the similar distribution. (2) More injected nodes would make the poisoning attack process noticeable. The results show that with the increase of $r$, the poisoned graph becomes more and more diverse from the origin graph. (3) The number of triangles increases, which shows that the attack not just simply connect fake nodes to other nodes, but also connect in a way to form triangles so each connection could affects more nodes. 

\subsection{Attack Effects Under Different Scenarios}
In this subsection, we conduct experiments to answer \textbf{RQ3}, i.e., how effective the attack by NIPA is under different scenarios.

\subsubsection{Average Degrees of Injected Nodes} \label{sec:degree_effect}
\begin{figure}[ht]
    \centering
    \includegraphics[width=8.5cm]{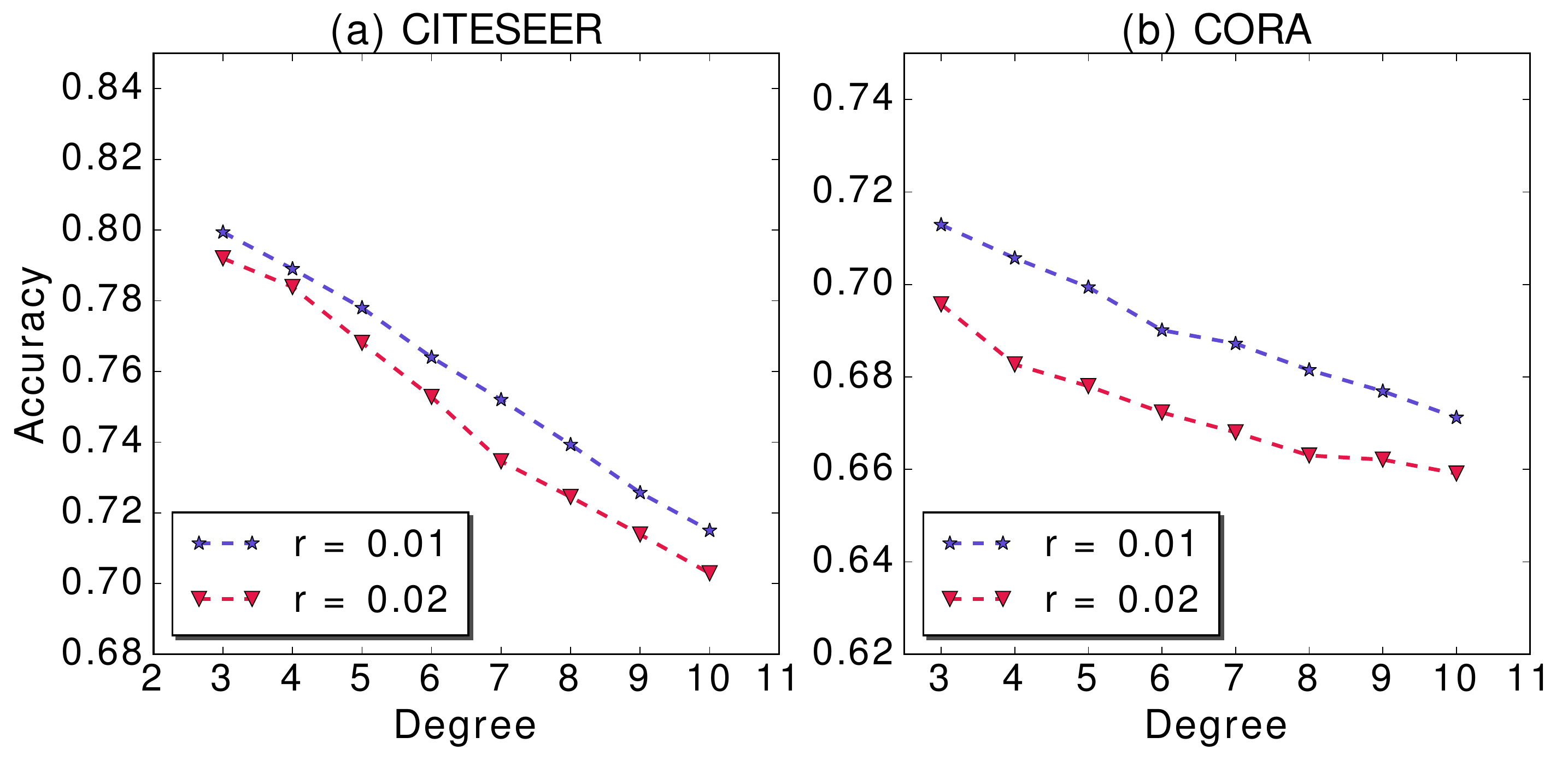}
    \vskip -1em
    \caption{Node classification performance on (a) CITESEER and (b) CORA by varying average node degree of injected nodes}
    \label{fig:degree}
    \vskip -1.5em
\end{figure}

As we discussed that the budget $\Delta=r*|V|*\text{deg}(v_\mathcal{A})$ is essential to the poisoning attack, we investigate the node classification accuracy by varying the average degree of injected nodes as $\text{deg}(v_\mathcal{A})=\{3,\dots 10\}$.
The experiment results when $r=0.1$ and $r=0.2$ on CITESEER and CORA are shown in Fig.~\ref{fig:degree}(a) and Fig.~\ref{fig:degree}(b), respectively. From the figures, we observe that as the increase of the average degree of the injected nodes, the node classification accuracy decrease sharply, which satisfies our expectation because the more links a fake node can have, the more likely it can poison the graph. 

\subsubsection{Sparsity of the Origin Graph}
\begin{figure}[h]
    \centering
   \includegraphics[width=8.5cm]{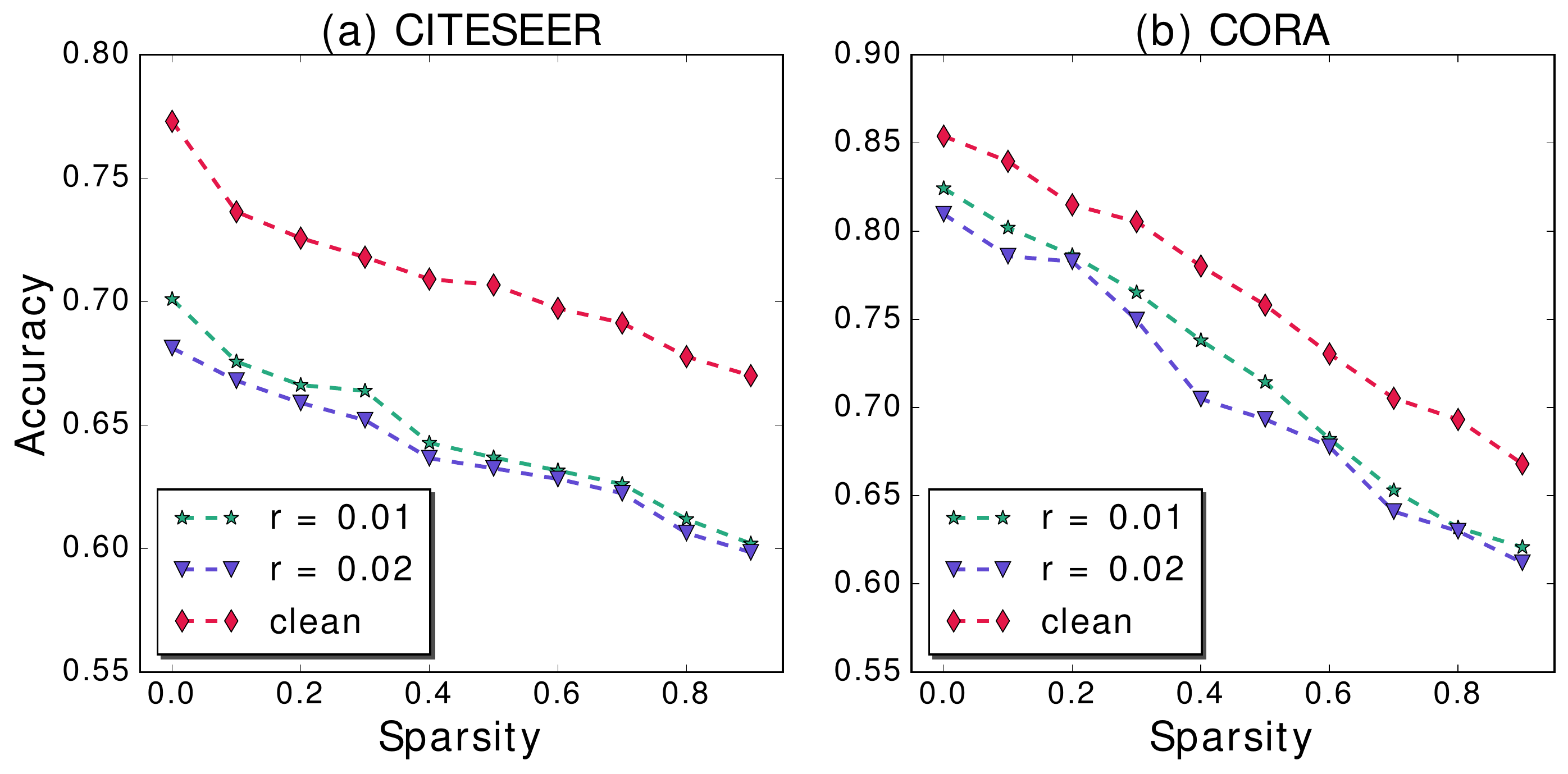}
   \vskip -1em
    \caption{Node classification performance on (a) CITESEER and (b) CORA with varying graph sparsity}
    \label{fig:sparsity}
    \vskip -1em
\end{figure}

We further investigate how the proposed framework works under different sparsity of the network. Without loss of generality, we set average degree of injected node as the average degree of the real node. To simulate the sparsity of the network, we randomly remove $S_p=\{0, 10\%, \dots, 90\%\}$ edges from the original graph. The results with $r=0.01$ and $r=0.02$ on CITSEER and CORA  are shown in Fig.\ref{fig:sparsity}. The results show that as the graph becomes more spare, the proposed framework is more effective in attacking the graph. This is because as the graph becomes more sparse, each node in the clean graph has less neighbors, which makes the it easier for fake nodes to change the labels of unlabeled nodes. 

\section{Conclusion}
In this paper, we study a novel problem of non-target graph poisoning attack via node injection. We propose \ourModelName a deep reinforcement learning based method to simulate the attack process and  manipulate the adversarial edges and labels for the injected nodes. Specifically, we design reward function and hierarchy DQNs to better communicate with the reinforcement learning environment and perform the poisoning attack. Experimental results on node classification demonstrate the effectiveness of the proposed framework for poisoning the graph. The poisoned graph has very similar properties as the original clean graph such as gini coefficient and distribution entropy. Further experiments are conducted to understand how the proposed framework works under different scenarios such as very sparse graph.

There are several interesting directions need further investigation. First, in this paper, we use mean values of node features as the feature of fake nodes. We would like to extend the proposed model for simultaneously generate features of fake nodes. Second, we would like to extend the proposed framework on more complicated graph data such as heterogeneous information network and dynamic network.


\end{document}